\documentclass{article}

\usepackage{PRIMEarxiv}
\usepackage{algorithm}
\usepackage{algpseudocode}
\usepackage{amsmath}
\usepackage{bbm}
\usepackage[utf8]{inputenc} 
\usepackage[T1]{fontenc}    
\usepackage{hyperref}       
\usepackage{url}            
\usepackage{booktabs}       
\usepackage{amsfonts}       
\usepackage{nicefrac}       
\usepackage{microtype}      
\usepackage{mathtools}
\usepackage{lipsum}
\usepackage{color}
\usepackage{tikz}
\usepackage{fancyhdr}       
\usepackage{graphicx}       
\usepackage{caption}
\usepackage{multirow}
\usepackage{float}
\graphicspath{{media/}}     

\DeclareMathOperator*{\argmin}{arg\,min}
\DeclareMathOperator*{\argmax}{arg\,max}

\usetikzlibrary{shapes,decorations,arrows,calc,arrows.meta,fit,positioning}
\tikzset{
    -Latex,auto,node distance =1 cm and 1 cm,semithick,
    state/.style ={ellipse, draw, minimum width = 0.7 cm},
    point/.style = {circle, draw, inner sep=0.04cm,fill,node contents={}},
    bidirected/.style={Latex-Latex,dashed},
    el/.style = {inner sep=2pt, align=left, sloped}
}

\title{Partial Counterfactual Identification for infinite Horizon Partially Observable Markov Decision Process}

\author{
  Aditya Kelvianto Sidharta \\
  Columbia University \\
  \texttt{aks2266@columbia.edu} \\
}

\begin{document}
\maketitle


\section*{Abstract}
This paper investigates the problem of bounding possible output from a counterfactual query given a set of observational data. While various works of literature have described methodologies to generate efficient algorithms that provide an optimal bound for the counterfactual query, all of them assume a finite-horizon causal diagram. This paper aims to extend the previous work by modifying Q-learning algorithm to provide informative bounds of a causal query given an infinite-horizon causal diagram. Through simulations, our algorithms are proven to perform better compared to existing algorithm. 

\section{Introduction}
A causal diagram is a representation of the existence or possible existence of relationships between different variables \cite{pearl1995causal}. The diagram takes the form of a Directed Acyclic Graph (DAG). An arrow within a DAG represents a potential direct relationship between one variable and the other, and bi-directed arrow represents a spurious relationship between two variables due to an exogenous (unobserved) variable. Accompanied by the method that is developed by Pearl, \textit{do-calculus}, we can utilize the causal diagram to determine whether a counterfactual query can be identified and estimated through its observational data. A general identification algorithm that relies on the decomposition of components based on the underlying Structural Causal Model has also been developed to provide a more systematic way to identify causal queries. \cite{tian2002general} These algorithms are proven to be sound and complete.

Cases where the algorithm fails to find admissible sets to identify the causal query are called \textit{non-identifiable} cases. In other words, based on the assumed causal graph, the causal query could not be estimated from the observational data. In these non-identifiable cases, there are several different parameterizations of the underlying Structural Causal Model that induces similar joint distribution of observed variables.\cite{pearl2000models}. As we do not have access to the underlying Structural Causal Model that induces the probability distribution that we observe in a real-world setting, we will not be able to get the exact point estimation of a causal query. 

Nevertheless, one can still utilize the observational data to construct an informative bound of its target probability given a causal query. Such an algorithm aims to constrain the set of possible values of the causal query output to be smaller than its original output domain. There are several bounding algorithms that have been developed to solve this issue. Earlier works utilize Instrumental Variables (IV) to provide non-parametric assumptions in order to provide informative bounds on the causal query \cite{manski1990nonparametric}. Canonical Partition has since been introduced to provide an algorithmic way to derive the causal bound systematically \cite{balke1994counterfactual}. The bound generated has been proven optimal for discrete and finite states. \textit{Canonical Structural Causal Models} has been developed to formalize the representation of such cases, and Monte Carlo algorithms are used to improve the speed of convergence of the bound estimation process. \cite{zhang2021partial} However, all of the aforementioned literature assumed finite-horizon states. This paper aims to provide an algorithm to estimate the partial identification bound for infinite-horizon states.

Markov Decision Process (MDP) formally describes the environment we use to perform reinforcement learning. The Markov Decision Process contains various states with corresponding transition dynamics that follow \textit{Markovian Property}. The Markov Decision Process also contains a reward function that indicates the time-step reward of performing certain actions. In this paper, we are concerned with infinite-horizon MDP, and thus we require the return to be multiplied against a time-discounted factor to ensure the total reward to be finite. \cite{howard1960dynamic}. A special case of MDP used in this paper is POMDP, \textit{Partially Observable Markov Decision Process}, where the agent interacting with the Markov Decision Process is not able to observe the underlying state directly. \cite{aastrom1965optimal} Formal definition of the environment will be defined in the \textbf{Preliminary} section.

Reinforcement Learning is a framework that allows a learning agent to collect information based on its own interaction or other's interaction to improve its' own performance on subsequent run \cite{sutton2018reinforcement}. One type of reinforcement learning that we are interested in exploring in this paper is model-free reinforcement learning, where the transition probability distribution of the associated Markov Decision Process is unknown. Various methods have been developed to tackle this problem, including Q-learning. Q-learning is an off-policy algorithm, meaning that it learns the optimal policy from a different behavioural policy. It performs policy evaluation using Bellman optimal equation and policy improvement based on its predicted Q-values. \cite{watkins1992q}. 

Q-learning assumes that the intervention data produced by the behaviour policies are induced from the same Structural Causal Model as the intervention distribution of our interest. In the case where we only have access to observational data, several literature have discussed methods to perform Q-learning in those situations. Methods like Propensity Scoring \cite{rosenbaum1983central}, Inverse Propensity of Treatment Weighting \cite{rotnitzky2005inverse}, and Covariate Adjustment \cite{austin2011introduction} has been used to tackle this issue. However, all of these methods require the strong ignorability assumption. In other words, an admissible set of covariates must be measured and included within the model such that the causal treatment effect can be estimated through adjustment. Thus, the above methods does not work in non-identifiable settings.

This paper aims to develop a novel algorithm based on Q-learning to perform Partial Identification for Infinite POMDP. We show that by defining a new set of Quality scores, ${Q_l}$ and ${Q_h}$, we can adapt Q-Learning to estimate the causal bound. We will first formally define the problem statement of this paper in the \textit{Preliminaries} section. Then, we will specify a novel Q-learning based algorithm to tackle this problem, which we call \textit{Partial Bound Q-learning}. Lastly, we show the results of our simulation that validates the result of our novel algorithm and prove that ignoring the non-identifiability of the causal graph will result in erroneous result. Finally, we will discuss potential future works on this problem.

\section{Preliminaries}

\subsection{Partially Observable Markov Decision Process}

Partially Observable Markov Decision Process can be represented with $ <S, X, Y, \Omega, T, R, O> $.

$S$ defines all of the possible underlying states of the agent. $X$ defines all of the possible actions of the agent. $Y$ defines the reward domain of the agent. $\Omega$ defines all of the observable states seen by the agent.

$T$ defines the transition distribution between the states. This transition distribution can be represented as a matrix, where the $(i, j)$-th entry of the matrix indicates the transition probability $T_{i}^{j} = p(S^{(t+1)} = j| S^{(t)} = i, do(X^{(t)} = x))$. The transition distribution can be represented as a matrix with shape $N \times N$, where $N$ is the number of states in the environment. Denote that Markovian property holds for POMDP - the probability of transition of the next state is assumed to be dependent only on the current state and the current action. In other words, given the current state and action, it is independent against historical states and actions. 

$R$ indicates the reward function, $R^{y}_{x, s} = E[Y^{(t)} = y | do(X^{(t)} = x], S^{(t)} = s)$. In cases where the output variable $Y^{(t)}$ is discrete, without loss of generality, the reward function can be represented as $R^{(t)} = \mathbbm{1}(Y^{(t)} = y) \; P(Y^{(t)} = y | do(X^{(t)} = x), S^{(t)} = s)$, where $\mathbbm{1}(x) = 1$ if $x$ is true and $0$ otherwise. 

$O$ indicate the conditional observation function, where the probability of observing observation state $o$ depends upon the underlying current state and the previous action taken. Concretely, $O_{s', x}^o = p(\Omega^{(t+1)} = o | S^{(t+1)} = s', do(x^{(t)} = x))$.

\subsection{Causal Model}

Structural Causal Model \cite{pearl2000models} is used as the semantical framework of our analysis. An SCM model M consists of 4 different components.

\begin{enumerate}
    \item $U$ is an exogenous variable. This component can affect the observable state ($\Omega$), reward state ($Y$), action state ($X$), and the underlying state ($S$) in our \textit{POMDP}.
    \item  $V$ is defined as endogenous (observed) variable. In our POMDP, this corresponds to our observable state $\Omega$, our action state $X$, and the reward state $Y$.
    \item $P(U)$ is the underlying distribution for each of our exogenous variable $U_i$, where each instance of the exogenous variable $U_i^{(t)}$ is sampled from its corresponding probability distribution $P(U_i)$ 
    \item F is a structural function that defines the values of each endogenous variable in our system, $V_i$, as a function of other endogenous and exogenous variables, i.e
    \begin{equation}
    V_i = f_i(Pa(V_i), U_i), \; \forall Pa(V_i) \in V \backslash V_i, U_i \in U
\end{equation}
\end{enumerate}

The goal of this paper is to bound a causal query in an infinite-horizon state, based on the observational data. In Structural Causal Model (SCM) setting, $do()$ operator is used to denote intervention. Therefore, in our conditional probabilities, given an action $x$, a notation $do(X^{(t)} = x)$ indicates that the agent manipulate the underlying model in setting the value of $X^{(t)}$ to be $x$ directly. In other words, $X^{(t)}$ is no longer induced by the parent variable and exogenous variable, $Pa(X^{(t)}), U_{X^{(t)}}$. In contrast, the notation $X^{(t)} = x$ without the $do()$ operator indicates that the action / treatment variable that we see is from observational data (not interventional), and the $X^{(t)}$ value is set through the underlying SCM. In other words, its the natural behavior that we observe in a real world setting, where the agent does not intervene on the system.

The Structural Causal Model can be extended to accommodate infinite-horizon setting, defined as Dynamic Structural Causal Model (DSCM) \cite{zhang2021can}. In this setting, The unobserved variable at time $t$ is dependent upon the previous action $x^{(t-1)}$, state $\Omega^{(t-1)}$, and unobserved variable $u^{(t-1)}$, as well as the corresponding noise $\epsilon^{(t)}_u$, i.e $u^{(t)} = f_u(x^{(t-1)}, \Omega^{(t-1)}, u^{(t-1)}, \epsilon^{(t)}_u)$. Similarly, the observable state can be defined as $\Omega^{(t)} = f_\Omega(x^{(t-1)}, s^{(t-1)}, u^{(t-1)}, u^{(t)})$. The Observed reward $y^{(t)}$ is defined as $y^{(t)} = f_y(x^{(t)}, s^{(t)}, u^{(t)})$. Lastly, the action state is defined as  $x^{(t)} = f_x(s^{(t)}, u^{(t)})$. Using this DSCM definition, we are able to use our identification and bounding method for infinite-state horizon POMDP.

\subsection{Partial Identification}

Let the probability distribution that is induced by an SCM M as $P_M(\cdot)$. The causal effect of $X$ on $Y$ is identifiable \cite{pearl2000models} from $G$ if and only if 

\begin{equation}
    P_{M_1}(y|do(x)) = P_{M_2}(y|do(x)) \; \forall \; {M_1, M_2 \in \{M | P_{M_1}(v) = P_{M_2}(v) > 0\}}
\end{equation}

Therefore, given a causal graph, a causal query $P(y|do(x))$ is non-identifiable if there exists at least one pair of SCM, corresponding to the given causal graph, that induces the same observational distribution $P_{M1}(v) = P_{M2}(v)$ but it has different interventional distribution $P_{M1}(y|do(x)) \neq P_{M2}(y|do(x))$.

The goal of \textit{partial-identification} algorithm is to bound the causal query of interest, $p(y|do(x))$ using $p(v)$ to a range of possible values that are smaller than it's original domain, i.e $P(y|do(x)) \in [a, b] \; s.t \; [a, b] \subset [0, 1]$. The lower bound and the upper bound of the causal query probability corresponds to instantiating SCM $M_{low}$ and $M_{high}$ such that $M_{low}$ is a Structural Causal Model that induces the minimum interventional probability and $M_{high}$ induces the maximum interventional probability while both $M_{low}$ and $M_{high}$ induces equal observational distribution $P(v)$, i.e 

\begin{equation}
    a = \min_{M_{low} \in M'} p(y|do(x)), \; b = \max_{M_{high} \in M'} p(y|do(x)), \; \forall M' \in \{ M | P_M'(v) = P_M(v)\}
\end{equation}

For any SCM $M'$ that is compatible with the following structural function:

\begin{equation}
\begin{gathered}
        z = f_z(u_{xz}, u_{yz})\\
        x = f_x(z, u_{xz}, u_{xy})\\
        y = f_y(y, u_{xy}, u_{yz})
\end{gathered}
\end{equation}

Given that we know the observation distribution $p(x,y,z)$, the Natural Bound of the causal query p(y|do(x), z) is bounded in [a, b] where\cite{manski1990nonparametric, balke1994counterfactual}:

\begin{equation}
    \begin{gathered}
        a = p(y,x|z) \\
        b = 1 + p(y,x|z) - p(x|z)
    \end{gathered}
\end{equation}

This natural bound assumption has been proven to be tight whenever the monotonicity assumption holds, i.e $p(x=x_1 | z=z_1, u) \geq p(x=x_1 | z=z_0, u)$ \cite{balke1997bounds}. 

\subsection{Q-Learning}
The goal of Q-learning algorithm is to learn an optimal policy $\pi^*(x|s) = p(X^{(t)} = x | S^{(t)} = s) \; \forall t$ through the trajectory obtained from a behavioral policy $\pi_b$  by estimating the value of an action at a particular state, $Q(s, x)$ \cite{watkins1992q}. Formally, Q-value is defined as the expected sum of rewards obtained when the agent follow the policy $\pi$ from being in a current state $S^{(t)} = s$ and taking action $X^{(t)} = x$, i,e:

\begin{equation}
    Q_\pi(S^{(t)} = s, X^{(t)} = x) = E_\pi [G^{(t)} | do(X^{(t)} = x), S^{(t)} = s], \text{ where } G^{(\tau)} = \sum_{t = \tau}^{T} \gamma^t (R^{(t)} = r)
\end{equation}

\begin{equation}
    Q_\pi(S^{(t)} = s, X^{(t)} = x) = E_\pi [r^{(t)} + \gamma (Q_\pi(S^{(t+1)} = s', X^{(t+1)} = x')) | do(X^{(t)} = x), S^{(t)} = s]
\end{equation}

\begin{equation}
    Q_\pi(S^{(t)} = s, X^{(t)} = x) = r^{(t)} + \gamma \sum_{s' \in S} T_{s,x}^{s'} \sum_{x' \in X} \pi(do(X^{(t+1)} = x')| S^{(t+1)} = s') Q_\pi(S^{(t+1)} = s', X^{(t+1)} = x')
\end{equation}

In other words, the Q-values can be defined as Bellman Expectation Equation, as shown in Equation [8], as it expresses relationship the the value function of an action-state pair in the current time step and the following time step.\cite{irodova2005reinforcement}. 

Q-learning utilizes the Bellman equation defined above to improve its policy evaluation, $Q_\pi(S^{(t)} = s, X^{(t)} = x)$. Specifically, it combines dynamic programming and Monte Carlo methods to solve the above Bellman equation defined above to obtain the optimal value function. It does that by updating the current prediction of the state-action pair iteratively using \textit{TD-error}, the difference between value function estimate of current state-action pair and the sum of reward and maximum value function estimate of the next state-action pair amongst all action.

\begin{equation}
    Q(s, x) \leftarrow  Q(S^{(t)} = s, X^{(t)} = x) + \alpha[(r^{(t)} + \gamma \max_{x'} Q(S^{(t+1)} = s', X^{(t+1)} = x')) - Q(S^{(t)} = s, X^{(t)} = x)]
\end{equation}

Using the trajectory sampled from the behavior policy, the transition probability from the current state to the next state is estimated through the frequency of appearance in the trajectory samples. Q-learning is a deterministic policy, and thus the next action taken by the policy will be the action that maximizes the Q-value of the next state. This determinism is the reason why updating our current state-action pair Q-value using \textit{TD-error} is valid, as the Q-learning equation is equivalent to the Bellman equation of a Q-value given in equation [7].

The output policy of Q-learning is a deterministic policy that chooses action which maximizes Q-values given a current state. Assuming that all state-action pairs are visited and updated by the behavior policy, Q-learning is guaranteed to converge to the optimal value $Q*$ \cite{sutton2018reinforcement}. Therefore, the policy that chooses action which maximizes the Q-values are guaranteed to be optimal, i.e

\begin{equation}
    \pi^*(S^{(t)} = s, X^{(t)} = x) = 1 \text{ if } x = \argmax_x Q^*(S^{(t)} = s, X^{(t)} = x), \text{0 otherwise}
\end{equation}

Q-learning algorithm is defined in the figure below:

\begin{algorithm}[H]
\caption{Q-learning}\label{alg:cap}
\begin{algorithmic}
\Procedure{Q-learning}{Dataset $\mathfrak{D}$, Number of Epoch $E$, States $S$, Actions $X$, Learning Rate $\alpha$, Discount Factor $\gamma$}
	\State Initialize $Q : |S| \times |X|$
	\For{$e = 1,...,E$}
		\For{$ d = 1,...,|D|$}
			\State $(s, x, r, s') = \mathfrak{D}[d]$
			\State $Q[s, x] = Q[s, x] + \alpha ((r + \gamma \max_{x'}(Q[s', x'])) - Q[s, x])$
    	\EndFor
	\EndFor
\State \Return $Q$
\EndProcedure
\end{algorithmic}
\end{algorithm}

Nevertheless, as discussed in the previous chapter, Q-learning requires strong ignorability assumption. In cases where Q-learning is used in non-identifiable causal graphs, the predicted Q-values will not be accurate. This will be shown in our \textbf{Experimentation} section below. 

In the next chapter, we will introduce \textit{Partial Bound Q-learning} to mitigate this issue.

\section{Partial Identification for Infinite State}

\subsection{Partial Bound Q-Learning}

\textit{Partial Bound Q-Learning} is an algorithm that aims to estimate the causal bound of a state-action pair in an infinite time horizon, given that our causal query is non-identifiable. In other words, we are interested to estimate $Q_h(S^{(t)} = s, X^{(t)}=x)$ and $Q_l(S^{(t)} = s, X^{(t)}=x)$, $\forall \; s \in S , x \in X , \; t \in T$. We will use $Q_h$ and $Q_l$ as a short form of the above notation throughout this paper. Using our \textit{Partial Bound Q-learning} algorithm, the $Q_h$ and $Q_l$ value that we estimate will represent the true causal bound of the non-identifiable DSCM, $[\tilde{Q}_l^{*}, \tilde{Q}_h^{*}]$, 

$G_{low}^{\pi}$ and $G_{high}^{\pi}$ is the the episodic return of the total discounted reward of policy $\pi$ induced by two different Dynamic Structural Causal Models that induce the same observational distribution,  $P(Y^{(t)}, X^{(t)} ,S^{(t)})$, and induce the lowest and highest expected interventional distribution respectively. We can formally define $G_{low}^{\pi}$ and $G_{high}^{\pi}$ as follows

\begin{equation}
\begin{gathered}
G_{low}^{\pi} = \sum_{t=0}^{t=T} \gamma^t \mathbbm{1}_{M_1}(y^{(t)} = y | do(X^{(t)}=x),S^{(t)} = s)\\
\\
G_{high}^{\pi} = \sum_{t=0}^{t=T} \gamma^t \mathbbm{1}_{M_2}(y^{(t)} =y | do(X^{(t)}=x),S^{(t)} = s)\\
\\
\text{where }p_{M_1}(Y^{(t)}, X^{(t)} ,S^{(t)}) = p_{M_2}(Y^{(t)}, X^{(t)} ,S^{(t)}) = p_{M'}(Y^{(t)}, X^{(t)} ,S^{(t)}), \; \forall M' \in M_{obs}\text{, }\\
\\
\text{and }M_1 = \argmin_{M_1 \in M_{obs}} E[\sum_{t=0}^{t=T} \gamma^t p_{M_1}(y^{(t)} = y | do(X^{(t)}=x),S^{(t)} = s)]\text{, }\\
\\
\text{and }M_2 = \argmax_{M_2 \in M_{obs}} E[\sum_{t=0}^{t=T} \gamma^t p_{M_2}(y^{(t)} = y |  do(X^{(t)}=x),S^{(t)} = s)]
\end{gathered}
\end{equation}

$\tilde{Q}_l^{\pi}, \tilde{Q}_h^{\pi}$ is the expected of the episodic return by following policy $\pi$ given the current state-action pair of the agent. 

\begin{equation}
\tilde{Q}^{\pi}_l(S^{(t)} = s, X^{(t)}=x) = E[G_{low}^{\pi} | do(X^{(t)}=x),S^{(t)} = s)]
\end{equation}

\begin{equation}
\tilde{Q}^{\pi}_h(S^{(t)} = s, X^{(t)}=x) = E[G_{high}^{\pi} | do(X^{(t)}=x),S^{(t)} = s)]
\end{equation}

The Bellman Equation of $\tilde{Q}^{\pi}_l(S^{(t)} = s, X^{(t)} = x)$ and $\tilde{Q}^{\pi}_h(S^{(t)} = s, X^{(t)} = x)$ follows similar structure as the original Q values. Thus, it can be defined as the following

\begin{equation}
    \tilde{Q}^{\pi}_l(S^{(t)} = s, X^{(t)} = x) = E_\pi [\tilde{a}^{(t)}(s, x) + \gamma \tilde{Q}^{\pi}_l(S^{(t+1)} = s', X^{(t+1)} = x') | do(X^{(t)} = x), S^{(t)} = s]
\end{equation}

\begin{equation}
    \tilde{Q}^{\pi}_h(S^{(t)} = s, X^{(t)} = x) = E_\pi [\tilde{b}^{(t)}(s, x) + \gamma \tilde{Q}^{\pi}_h(S^{(t+1)} = s', X^{(t+1)} = x') | do(X^{(t)} = x), S^{(t)} = s]
\end{equation}

$\tilde{a}$ and $\tilde{b}$ is the lower and higher bound reward of policy $\pi$ at a single time-step $t$ respectively, i.e

\begin{equation}
    \begin{gathered}
    \tilde{a}^{(t)}(s, x) = p_{M_1}(y^{(t)} = y | do(X^{(t)}=x),S^{(t)} = s)\\
    \tilde{b}^{(t)}(s, x) = p_{M_2}(y^{(t)} = y |  do(X^{(t)}=x),S^{(t)} = s)
    \end{gathered}
\end{equation}

Based on derivation in \cite{manski1990nonparametric}, we can bound the causal query as follows:

\begin{equation}
    \begin{split}
        p(Y^{(t)} = y | do(X^{(t)} = x), S^{(t)} = s) & = \frac{p(Y^{(t)} = y, do(X^{(t)} = x), S^{(t)} = s)}{p(do(X^{(t)} = x), S^{(t)} = s)} \\
        & = \frac{\sum_u p(Y^{(t)} = y| X^{(t)} = x, S^{(t)} = s, U^{(t)} = u)p(S^{(t)} = s)p(U^{(t)} = u)}{p(S^{(t)} = s)} \\ 
        & = \sum_u p(Y^{(t)} = y| X^{(t)} = x, S^{(t)} = s, U^{(t)} = u)p(U^{(t)} = u) \\
        & = \begin{multlined}[t]
           \sum_u p(Y^{(t)} = y| X^{(t)} = x, S^{(t)} = s, U^{(t)} = u)\\(p(X^{(t)} = x, U^{(t)} = u | S^{(t)} = s) - p(X^{(t)} = x, U^{(t)} = u | S^{(t)} = s) \\ + p(U^{(t)} = u))
     \end{multlined} \\
       & = \begin{multlined}[t]
       p(Y^{(t)} = y, X^{(t)} = x | S^{(t)} = s) + \\ \sum_u p(Y^{(t)} = y | X^{(t)} = x, U^{(t)} = u, S^{(t)} = s)\\(p(U^{(t)} = u) - p(X^{(t)} = x, U^{(t)} = u | S^{(t)} = s)) 
     \end{multlined} \\
      & \geq p(Y^{(t)} = y, X^{(t)} = x | S^{(t)} = s)
    \end{split}
\end{equation}

\pagebreak

This is true because conditional probability of $p(y|x,u,s) \geq 0$. Therefore, Equation [17] forms the lower bound of the natural bound for the intervention query.

\begin{equation}
    \tilde{a}^{(t)}(s, x) = p(Y^{(t)} = y, X^{(t)} = x | S^{(t)} = s)
\end{equation}

Using the similar equation, the upper bound can be derived as follows

\begin{equation}
    p(Y^{(t)} = y | do(X^{(t)} = x), S^{(t)} = s) \leq p(Y^{(t)} = y, X^{(t)} = x | S^{(t)} = s) + 1 - p(X^{(t)} = x | S^{(t)} = s)
\end{equation}

This is true because conditional probability of $p(y|x,u,s) \leq 1$. Therefore, 

\begin{equation}
    \tilde{b}^{(t)}(s, x) = p(Y^{(t)} = y, X^{(t)} = x | S^{(t)} = s) + 1 - p(X^{(t)} = x | S^{(t)} = s)
\end{equation}

\textit{Partial Bound Q-Learning} uses $B$ records for each epoch $E$ in order to calculate the estimated bound $\hat{a}(s, x)$ and $\hat{b}(s, x)$. The number of matching tuples within a batch $n = \sum_{i \in \mathfrak{B}} \mathbbm{1} (S_i^{(t)} = s, S_i^{(t+1)} = s')$ will be used as a multiplier to learning rate $\alpha$ as unbiased Monte Carlo estimate of the transition probability in model-free environment. Therefore, in the case where $B$ is equal to 1, the \textit{Partial Bound Q-Learning} operates similarly to the vanilla Q-learning algorithm, updating the Q-value for each records it receive.

To be specific, the lower bound and the upper bound value would be calculated empirically using the trajectory from the observational data. we are able to estimate the lower bound and higher bound, $\hat{a}(s, x)$ and $\hat{b}(s, x)$, as follows

\begin{equation}
    \begin{gathered}
    \hat{a}(s, x) = \frac{\sum_{i\in B, t \in T} \mathbbm{1}(y^{(t)}_i = 1,  X^{(t)}_i=x ,S^{(t)}_i = s)}{\sum_{i\in B, t \in T} \mathbbm{1}(S^{(t)}_i = s)}\\
    \\
    \hat{b}(s, x) = 1 + \frac{\sum_{i\in B, t \in T} \mathbbm{1}(y^{(t)}_i = 1,  X^{(t)}_i=x ,S^{(t)}_i = s)}{\sum_{i\in B, t \in T} \mathbbm{1}(S^{(t)}_i = s)} - 
    \frac{\sum_{i\in B, t \in T} \mathbbm{1}( X^{(t)}_i=x ,S^{(t)}_i = s)}{\sum_{i\in B, t \in T} \mathbbm{1}(S^{(t)}_i = s)}
    \end{gathered}
\end{equation}

In the proposed implementation below, we use tabular Q-learning method to store and update the values of $Q_l$ and $Q_h$. In other words, for all unique state and action the agent can take, we will initialize value of $Q_l(s, x)$ and $Q_h(s, x)$ and perform the update on the corresponding values iteratively. Similar to Deep Q-learning, the tabular method for \textit{Partial Bound Q-learning} can be swapped with any other value approximation function (i.e Neural Network), in order to deal with environment with large state domain.\cite{mnih2013playing}

\pagebreak

In order to perform policy evaluation, similar to the Q-learning algorithm, the proposed algorithm uses bootstrapping method, where it utilizes the Bellman Equation of the value function to perform the value iteration update based on the observed reward value and current value estimation of the next state. 

\begin{equation}
    \begin{gathered}
    Q_l[s, x] = Q_l[s, x] + n\alpha ((\hat{a} + \gamma \max_{x'}(Q_l[s', x'])) - Q_l[s, x])\\
	Q_h[s, x] = Q_h[s, x] + n\alpha ((\hat{b} + \gamma \max_{x'}(Q_h[s', x'])) - Q_h[s, x])\\
	\text{where }n = \sum_{i \in \mathfrak{B}} \mathbbm{1} (S_i^{(t)} = s, S_i^{(t+1)} = s') 
    \end{gathered}
\end{equation}

The complete algorithm of \textit{Partial Bound Q-learning} is defined in the figure below:

\begin{algorithm}[H]
\caption{Partial Bound Q-learning}\label{alg:cap}
\begin{algorithmic}
\Procedure{PBQL}{Dataset $\mathfrak{D}$, Number of Epoch $E$, Batch size $B$, States $S$, Actions $X$, Learning Rate $\alpha$, Discount Factor $\gamma$}
	\State Initialize $Q_l : |S| \times |X|$
	\State Initialize $Q_h : |S| \times |X|$
	\For{$e = 1,...,E$}
		\For{$b = 1,...,B$}
    		\For{$ s = 1,...,|S|$}
    			\For{$ x = 1,...,|X|$}
    			        \State $\mathfrak{B} = \langle S^{(t)}_b = s, X^{(t)}_b = x, Y^{(t)}_b = y, S^{(t+1)}_b = s' \rangle \; \forall b \in [\frac{b|\mathfrak{D}|}{B}, \frac{(b+1)|\mathfrak{D}|}{B})$
        				\State $\hat{a} \approx p_\mathfrak{B}(Y^{(t)} = 1, X^{(t)}=x | S^{(t)} =  s), \forall \; t = (1, \infty)$
        				\State $\hat{b} \approx 1 + p_\mathfrak{B}(Y^{(t)} = 1, X^{(t)}=x |S^{(t)} = s) - p_\mathfrak{B}(X^{(t)}=x  | S^{(t)} = s), \forall \; t = (1, \infty)$
        				\For {$ s' = 1,...,|S|$}
        				    \State $n = \sum_{i \in \mathfrak{B}} \mathbbm{1} (S_i^{(t)} = s, S_i^{(t+1)} = s')$
        				    \If{$n > 0$}
                				\State $Q_l[s, x] = Q_l[s, x] + n\alpha ((\hat{a} + \gamma \max_{x'}(Q_l[s', x'])) - Q_l[s, x])$
                				\State $Q_h[s, x] = Q_h[s, x] + n\alpha ((\hat{b} + \gamma \max_{x'}(Q_h[s', x'])) - Q_h[s, x])$	
                			\EndIf
        			\EndFor
    			\EndFor
    		\EndFor
    	\EndFor
	\EndFor
\State \Return $Q_l$, $Q_h$
\EndProcedure
\end{algorithmic}
\end{algorithm}

\pagebreak

Unlike Q-learning, the policy learned from the \textit{Partial Bound Q-learning} method would be a stochastic policy, where the probability of playing action $x$ given that $S^{(t)} = s$ equals to the probability that it maximizes the expected reward $\tilde{y}$, assuming that $\tilde{y}$ is distributed uniformly between $Q_l[s, x]$ and $Q_h[s, x]$ out of all possible actions $x' \in X$. In other words, we utilize \textit{Thompson Sampling method} to implement the probability matching to get the probability of outputting certain actions. Formally, the stochastic policy of the \textit{Partial Bound Q-learning} algorithm is defined as follows

\begin{equation}
    \pi(X^{(t)} = x, S^{(t)} = s) = P(\tilde{Q}(x|s) > \tilde{Q}(x'|s), \forall \; x' \in X \backslash x),\; \tilde{Q}(\tilde{x}|s) \sim U(Q_l[s, \tilde{x}], Q_h[s, \tilde{x}])
\end{equation}

\begin{equation}
    \pi(X^{(t)} = x, S^{(t)} = s) = E_{[Q_l, Q_h]|\mathfrak{D}} [\mathbbm{1}(x = \argmax_{x \in X} \tilde{Q}(x|s))] 
\end{equation}

The planning algorithm used for the \textit{Partial-Bound Q-learning}  is given as follows

\begin{algorithm}[H]
\caption{Partial Bound Q-learning Planning}\label{alg:cap}
\begin{algorithmic}
\Procedure{PBQL-Plan}{Environment $\mathfrak{E}$, Lower Bound Q-value $Q_l[s, x]$, Upper Bound Q-value $Q_h[s, x]$, Discount factor $\gamma$}
	\State $s \sim S^{(t = 0)}$ \text{(current state)}
	\State $d = False$ \text{(terminate state reached)}
	\State $r = 0$ \text{(total discounted instance reward)}
	\State $t = 0$ \text{(time step)}
	\While{$d == False$}
	    \State $Q = -\epsilon, \;\epsilon >> 0$
	    \For{$\tilde{x} \in X$}
	        \State sample $\tilde{Q} \sim U(Q_l[s, \tilde{x}], Q_h[s, \tilde{x}])$
	        \If {$\tilde{Q} > Q$}
	            \State $x = \tilde{x}$
	            \State $Q = \tilde{Q}$
	       \EndIf
	    \EndFor
	    \State $y, s', d = \mathfrak{E}(s, x)$ \text{(time step reward, next state, terminate state reached)}
	    \State $r = r + y\gamma^{t}$
	    \State $t = t + 1$
	    \State $s = s'$
	\EndWhile
\State \Return $r$
\EndProcedure
\end{algorithmic}
\end{algorithm}

\section{Experiments}

\subsection{Example}

The motivating example used to illustrate our novel algorithm is modified from the original MDPUC causal diagram specified in \cite{zhang2021can}. In this setting, $S^{(t)} = s^{(t)}, s^{(t)} \in \{0,1\} \; \forall \; t \in T $ indicates the state that can be observed by the agent at time $t$. $X^{(t)} = x^{(t)}, x^{(t)} \in \{0,1\} \; \forall \; t \in T$ indicates the binary action / treatment taken by the behavior policy that induces the observational data $\pi_{obs}$. Based on the reward function of the environment, the agent will observe the output at time $t$, denoted by $Y^{(t)} = y^{(t)}, y^{(t)} \in \{0, 1\} \; \forall \; t \in T$. The current action, output, as well as the next state in the observational data, $X^{(t)}$, $Y^{(t)}$, and $S^{(t+1)}$, is also affected by unobserved (exogenous) variable $\; U^{(t)} = u^{(t)}, u^{(t)} \in \{0,1\} \; \forall \; t \in T$. This exogenous variable is not observed by the agent. The MDPUC is an infinite-horizon Partially Observed Markov Decision Process, where $t$ indicate the current time step of the agent. The MDPUC causal diagram is illustrated below in the figure below

\begin{tikzpicture}
    \node[state] (x1) at (0,0) {$X^{(1)}$};
    \node[state] (y1) at (3,0) {$Y^{(1)}$};
    \node[state] (s1) at (0,1.5) {$S^{(1)}$};
    \node[state] (u1) at (1.5,3) {$U^{(1)}$};
    
    \node[state] (x2) at (6,0) {$X^{(2)}$};
    \node[state] (y2) at (9,0) {$Y^{(2)}$};
    \node[state] (s2) at (6,1.5) {$S^{(2)}$};
    \node[state] (u2) at (7.5,3) {$U^{(2)}$};
    
    \node (s3) at (12, 1.5) {...};
    
    \node[state] (st) at (15,1.5) {$S^{(t)}$};
    
    \path (s1) edge (x1);
    \path (s1) edge (y1);
    \path (s1) edge (s2);
    \path (x1) edge (y1);
    \path (x1) edge (s2);
    \path[dotted] (u1) edge (x1);
    \path[dotted] (u1) edge (y1);
    \path[dotted] (u1) edge (s2);
    
    \path (s2) edge (x2);
    \path (s2) edge (y2);
    \path (s2) edge (s3);
    \path (x2) edge (y2);
    \path (x2) edge (s3);
    \path[dotted] (u2) edge (x2);
    \path[dotted] (u2) edge (y2);
    \path[dotted] (u2) edge (s3);
    
    \path[dotted] (s3) edge (st);

\end{tikzpicture}

In order to make the simulation concrete, we assume an example of a problematic clinical assessment of a dangerous drug. Assume that a new variant of non-lethal, long-term lung disease has recently swept the entire world and affect every single person. Unfortunately, the disease will reside in the body indefinitely ($t = (0, \infty)$). This lung disease may or may not cause symptoms to the patients. In other words, an infected patient might be asymptomatic ($s^{(t)} = 1$) even though one might suffer from the disease.

Suppose that a new type of drug has been recently developed to tackle the disease. However, due to the cost of research and production, the distribution of the drug has been unequal. All rich people ($u^{(t)} = 1$) are able to afford and obtain the drug ($x^{(t)} = 1$ if treatment is administered, $0$ otherwise), while only small percentage poor people ($u^{(t)} = 0$) are able to obtain it. The drug company has failed to take the inequality distribution of drug into account, and thus this is an unobserved variable.

Unknown to the drug company, the drug is actually \textit{detrimental} to the lung condition (healthy when $y^{(t)} = 1$, non-healthy otherwise) of a patient. Unbeknown to the drug manufacturer, beside obtaining this drug, all rich people has taken separate treatment that they took in conjunction, which neutered the effect of this drug, and actually improve the lung conditions. Thus, because of this separate treatment, all rich people are healthy are unaffected from this disease. All poor people do not have access to this other treatment. Therefore, people who obtained the drug and does not have access to the aforementioned treatment actually have worse lung condition compared to those who does not take the drug. Furthermore, once they took the drug, they will also be consistently symptomatic, and unable to recover from it.

The Markov Reward Process of the example MDPUC explained above is given as follows

\begin{table}[!htb]
\begin{tabular}{|l|l|}
\hline
p($u^{(t)} = 1$) & 0.25 \\ \hline
p($u^{(t)} = 0$) & 0.75 \\ \hline
\end{tabular}
\end{table}

\begin{table}[!htb]
\begin{tabular}{|l|l|}
\hline
p($s^{(1)} = 1$) & 0.50 \\ \hline
p($s^{(1)} = 0$) & 0.50 \\ \hline
\end{tabular}
\end{table}

\begin{table}[!htb]
\begin{tabular}{|l|l|l|}
\hline
p($x^{(t)} = 1$) & \multicolumn{1}{l|}{$s^{(t)} = 0$} & \multicolumn{1}{l|}{$s^{(t)} = 1$} \\ \hline
$u^{(t)} = 0$    & 0.10                              & 0.10                              \\ \hline
$u^{(t)} = 1$    & 1.00                              & 1.00                              \\ \hline
\end{tabular}
\end{table}

\begin{table}[!htb]
\begin{tabular}{|l|ll|ll|}
\hline
\multicolumn{1}{|c|}{\multirow{2}{*}{p($y^{(t)} = 1$)}} & \multicolumn{2}{c|}{$x^{(t)} = 0$}                 & \multicolumn{2}{c|}{$x^{(t)} = 1$}                 \\ \cline{2-5} 
\multicolumn{1}{|c|}{}                                  & \multicolumn{1}{l|}{$s^{(t)} = 0$} & $s^{(t)} = 1$ & \multicolumn{1}{l|}{$s^{(t)} = 0$} & $s^{(t)} = 1$ \\ \hline
$u^{(t)} = 0$                                           & \multicolumn{1}{l|}{0.25}           & 0.50           & \multicolumn{1}{l|}{0.00}           & 0.00           \\ \hline
$u^{(t)} = 1$                                           & \multicolumn{1}{l|}{1.00}           & 1.00           & \multicolumn{1}{l|}{1.00}           & 1.00           \\ \hline
\end{tabular}
\end{table}

\begin{table}[!htb]
\begin{tabular}{|l|ll|ll|}
\hline
\multicolumn{1}{|c|}{\multirow{2}{*}{p($s^{(t+1)} = 1$)}} & \multicolumn{2}{c|}{$x^{(t)} = 0$}                 & \multicolumn{2}{c|}{$x^{(t)} = 1$}                 \\ \cline{2-5} 
\multicolumn{1}{|c|}{}                                    & \multicolumn{1}{l|}{$s^{(t)} = 0$} & $s^{(t)} = 1$ & \multicolumn{1}{l|}{$s^{(t)} = 0$} & $s^{(t)} = 1$ \\ \hline
$u^{(t)} = 0$                                             & \multicolumn{1}{l|}{0.50}           & 0.50           & \multicolumn{1}{l|}{0.00}           & 0.00           \\ \hline
$u^{(t)} = 1$                                             & \multicolumn{1}{l|}{0.50}           & 0.50           & \multicolumn{1}{l|}{0.50}           & 0.50           \\ \hline
\end{tabular}
\end{table}

Therefore, the optimal policy for this infinite-horizon, MDPUC will not to administer the drug regardless of the state and time horizon, i.e $\pi^*(s^{(t)} = s, x^{(t)} = 0) = 1 \; \forall s \in \{0, 1\}, \; \forall t$.  However, the problem is non-trivial when we do not have access to the financial status of the individual and we are not aware about the inequality in the drug distribution. We will compare the result of \textit{Partial Bound Q-Learning} and the vanilla \textit{Q-Learning} on the simulation data based on the MDPUC described above.

\subsection{Methodology}

We will compare the performance of both \textit{Partial Bound Q-learning} and vanilla Q-learning that do not have access to the exogenous variable. However, while the \textit{Partial Bound Q-learning} is aware about the fact that the causal query is non-identifiable, the vanilla Q-learning is unaware about the existence of the confounding variable.

These two algorithms will be fitted on samples of observational data obtained from the MDPUC stated above. The goal of these algorithms are
\begin{enumerate}
    \item Predict or bound the optimal Q-values of the interventional distribution given the initial state and action.
    \item Predict the optimal action / intervention to maximize the reward based on the observed state from the environment.
\end{enumerate}

We will prove that while the \textit{Partial Bound Q-learning} is non-optimal due to the nature of non-identifiability query, it still performs better than vanilla Q-learning algorithm.

In generating the observational data samples, we create $N = 1000$ instances, each with $T=500$ time steps. The discount rate of reward is assumed to be $\gamma = 0.90$. The learning rate used in both algorithms is set to be $alpha = 0.05$. Both of the algorithms are trained with $E = 500$ epochs. For the \textit{Partial-Based Q-Learning} algorithm, we use $B=1$ as the batch size. In other words, the $Q_l[s,x], Q_h[s,x]$ is updated for every record, similar to the update process of a vanilla Q-learning algorithm 

\subsection{Result}

As discussed in the previous chapter, given that we know the underlying Dynamic Structural Causal Model of the MDPUC, The optimal policy of the MDPUC is $\pi^*(s^{(t)} = s, x^{(t)} = 0) = 1 \; \forall s \in \{0, 1\}$ for all timestep $t$. Therefore, we are able to calculate the optimal Q-values for both states $s^{(t)} \in \{0, 1\}$ and actions $x^{(t)} \in \{0, 1\}$. The calculation of optimal Q-value, given that we know the underlying SCM of MDPUC, is given below.

\begin{equation}
    \begin{split}
        Q^*[s, x] =  \sum_u R(s,x,u) + (\gamma \sum_{s'} (p(s^{(t+1)} = s' | x^{(t)} = x, s^{(t)} = s, u^{(t)} = u))(\frac{R(s=s', x=0, u=u)}{1 - \gamma}))
    \end{split}
\end{equation}

where $R(s, x, u) = p(y^{(t)} =  1 | x^{(t)} = x, s^{(t)} = s, u^{(t)} = u) p(u^{(t)} = u)$

\bigskip

Calculating these values for all permutations of states and actions, we get the following results

\begin{equation}
    \begin{gathered}
    Q[s=0, x=0] = 5.219 \\
    Q[s=0, x=1] = 5.031 \\
    Q[s=1, x=0] = 5.406 \\
    Q[s=1, x=1] = 5.031
    \end{gathered}
\end{equation}

The first experimentation that we did is to run the Vanilla Q-learning algorithm on the simulation data, without acknowledging the confounding variables that might affect the relationship between the treatment and outcome variables. After running the algorithm on the simulated data, the resulting Q-values is given as follows

\begin{table}[H]
\begin{tabular}{|l|l|l|}
\hline
\multicolumn{1}{|c|}{$Q-value$} & \multicolumn{1}{c|}{$x^{(t)} = 0$} & $x^{(t)} = 1$ \\ \hline
$s^{(t)} = 0$                 & 7.206                       & 7.637    \\ \hline
$s^{(t)} = 1$                 & 7.466                       & 7.703    \\ \hline
\end{tabular}
\end{table}

Unsurprisingly, as the observational data induced by the Dynamic Structural Causal Model causes the trial drug to be distributed amongst the wealthy people that are healthy and do not adversely affected by the drug, and the population that do not receive drug are the poor people that generally suffer from worse health conditions compared to the rich people. Therefore, solely learning from observational distribution without taking into account the confounding variables (wealth) will lead towards false conclusion that the drug has a positive effect towards the lung condition. In other words, the vanilla Q-learning algorithm will estimate that the Q-value of giving the drug, $Q[s, x=1]$ will be higher for all state $s \in \{0,1\}$. Moreover, as drugs are mostly administered on healthier group of population, the Q-values estimated from the observational data over-predict the true Q-values of the optimal intervention policy.

Next, we run our \textit{Partial Bound Q-learning Algorithm} on the simulated data. The algorithm is proven to be useful if the true Q-values of the MDPUC for all state-action pairs are contained within the bound of the resulting Q-values from the corresponding algorithm. After running the algorithm, resulting Q-value bounds is as follows

\begin{table}[H]
\begin{tabular}{|l|l|l|}
\hline
\multicolumn{1}{|c|}{$Q_{low}$} & \multicolumn{1}{c|}{$x^{(t)} = 0$} & $x^{(t)} = 1$ \\ \hline
$s^{(t)} = 0$                 & 2.602                         & 2.631    \\ \hline
$s^{(t)} = 1$                 & 2.760                         & 2.670    \\ \hline
\end{tabular}
\end{table}

\begin{table}[H]
\begin{tabular}{|l|l|l|}
\hline
\multicolumn{1}{|c|}{$Q_{high}$} & \multicolumn{1}{c|}{$x^{(t)} = 0$} & $x^{(t)} = 1$ \\ \hline
$s^{(t)} = 0$                 & 8.784                        & 9.233    \\ \hline
$s^{(t)} = 1$                 & 8.976                        & 9.234    \\ \hline
\end{tabular}
\end{table}

Based on the final result, we observe \textit{Partial Bound Q-learning} algorithm is able to predict an accurate bound that contain the true total rewards for each state-action pair. However, due to the \textit{bow-graph} structure within the causal graph, and the fact that we are working with an infinite horizon Markov chain, the interval of the bound has been multiplied exponentially across different timesteps. This causes the bound to be large. Nevertheless, the bound is still tight, given the assumed causal graph of the problem.

The source of both inaccuracies in vanilla Q-learning algorithm and the wide total reward bound in \textit{Partial Bound Q-learning} is due to lack of observational data in poor people that receive treatment $<U^{(t)} = 1, X^{(t)} = 1, Y^{(t)} = y>$. \textit{Partial Bound Q-learning} is still able to use this information to provide a valid total reward bound for the problem, while vanilla Q-learning is unable to take this into account

The trend of Q-values estimation for each iteration for both vanilla Q-learning and \textit{Partial Bound Q-learning} is given below

\begin{figure}[H]
\includegraphics[width=10cm]{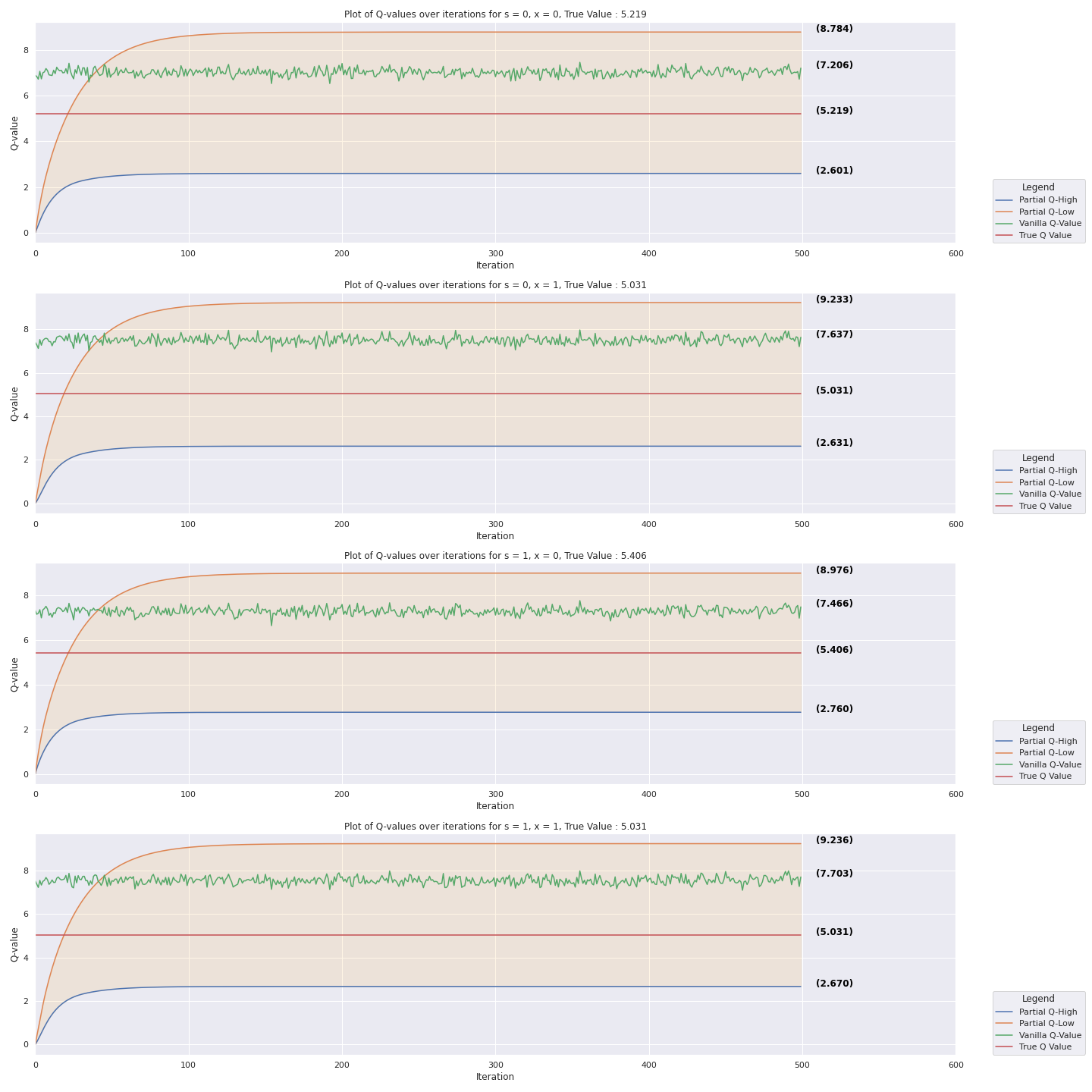}
\end{figure}

The Q-values observed from the \textit{Partial Bound Q-learning} and vanilla Q-learning algorithm is then used as a policy that interact with the simulation environment. An optimal policy is a policy that are able to maximize the total sum of discounted reward of the infinite-horizon MDPUC. Therefore, we are able to define \textit{regret} to compare the performance of various policies.

\textit{Regret} is defined as total loss of reward across the total horizon for the agent to follow a given policy instead of following the optimal policy. Therefore, a policy that accumulates higher regret means that it performs sub-optimally compared to the other policy.

\begin{equation}
    L_{\pi} = E[\sum_{t=1}^{\infty} \gamma^{t} (p(Y^{(t)} = y | do(\pi^*(S^{(t)} = s)), S^{(t)} = s) - (p(Y^{(t)} = y | do(\pi(S^{(t)} = s)), S^{(t)} = s))]
\end{equation}

In our data simulation, we will run the optimal policy $\pi^*$, \textit{Partial Bound Q-learning} policy $\pi^{PBQ}$, and vanilla Q-learning algorithm $\pi^{Q}$ against  $N = 5000$ instances. The histogram of each individual total instance rewards for performing intervention using the different policies is shown below.

\begin{figure}[H]
\includegraphics[width=15cm]{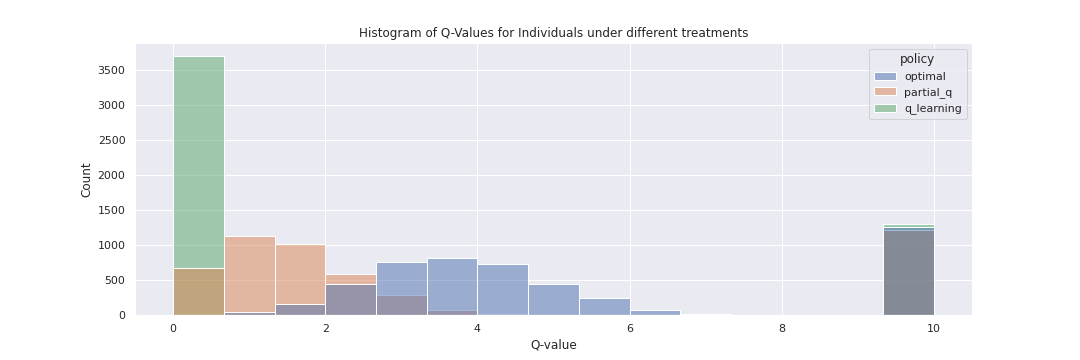}
\end{figure}

As we can see from the figure above, both the \textit{Partial Bound Q-learning} and vanilla Q-learning policy perform sub-optimally compared to the optimal policy. The gap in performance is expected due to the fact that we are dealing with a non-identifiable case. Nevertheless, we observe that the mean total instance reward under the \textit{Partial Bound Q-learning} is still significantly higher compared to the vanilla Q-learning algorithm. This is because the Q-learning algorithm mistakenly conclude that the Q-values for assigning treatment is higher than non-treatment for any given state. This policy has been proven to be detrimental to the poor people $(U^{(t)} = 0)$, and thus causing the mean total reward to be smaller.

Similarly, the regret of both \textit{Partial Bound Q-learning} and the vanilla Q-learning algorithm can be observed from the diagram below. The regret is the gap between the optimal policy curve and the corresponding policies in the figure below. The diagram shows the cumulative mean total reward of an instance across different time steps when we perform different policies.

\begin{figure}[H]
\includegraphics[width=15cm]{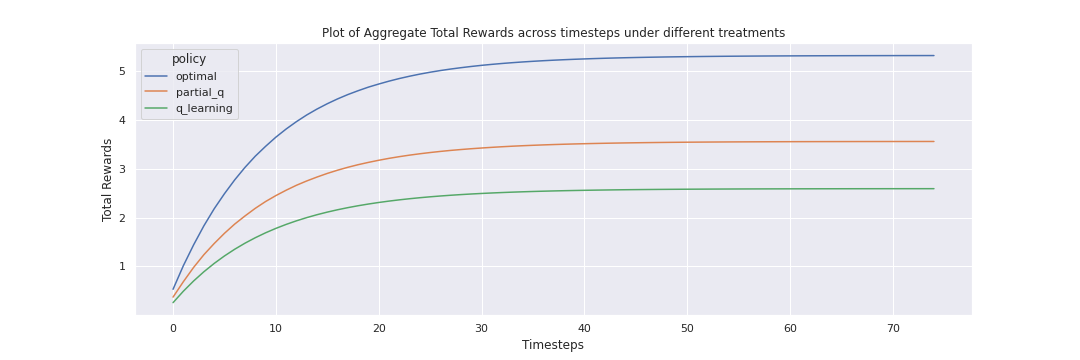}
\end{figure}

From the simulation, we have managed to prove that the \textit{Partial Bound Q Learning} is able to estimate the causal bound of a Dynamic Structural Causal Model. We have also proven that this causal bound is important for the policy to not be stuck choosing sub-optimal actions.

\section{Future Works}

Over the years, there has been various improvement that has been introduced to improve the convergence and scalability of Q-learning algorithm. Specifically, Deep Q Network \cite{mnih2013playing} has been introduced to replace the tabular Q-value with a value approximation function so that it is able to handle huge state-action space. The value function approximation might also be able to generalize from the earlier experiences to estimate the quality score of unseen states. In combining value function approximator with \textit{Partial Bound Q-learning}, we might employ target network in order to make the parameter update on the value function approximator stable.

In this paper, we assume that we are dealing with off-policy setting, where the agent only have access to observational data from a non-identifiable causal graph. In other words, the agent does not perform policy improvement based on it's own intervention. In the case where the agents are able to interact with the environment, the partial bound estimation can be further tightened using the intervention data. Bayesian approaches can be used to combine the prior causal bound obtained with the intervention data from the agents interaction. This method will allow us to create a Generalized Policy Learning (GPL) that works for infinite-horizon MDP and non-identifiable causal query.

\section{Conclusion}

In this paper, we have introduced a novel algorithm to bound a possible output from a counterfactual query in infinite-horizon, non-identifiable POMDP. In particular, we have modified Q-learning algorithm in order to provide a bounded Q-values of our causal query. We show that the information that we collect from non-identifiable casual graph and observational data are useful for us to generate policy that performs better than vanilla Q-learning algorithm. We have also discussed the potential future works on this topic.

\bibliographystyle{unsrt}  
\bibliography{references}  

\end{document}